\documentclass[letterpaper, 10 pt, conference]{ieeeconf}  

\IEEEoverridecommandlockouts                              

\makeatletter
\let\NAT@parse\undefined
\makeatother
\usepackage[numbers]{natbib}

\usepackage[pdftex]{graphicx}
\usepackage{amsmath}
\usepackage{amssymb}  
\usepackage{subfigure}
\usepackage{multirow}
\usepackage{array,booktabs}
\usepackage{diagbox}
\usepackage{balance}
\usepackage[ruled]{algorithm}
\usepackage{algpseudocode}
\newcolumntype{P}[1]{>{\centering\arraybackslash}p{#1}}
\usepackage{bm}
\newcommand*\boldell{\pmb{\ell}}

 \usepackage[final]{hyperref}
 \hypersetup{
     colorlinks=true,
     linkcolor=blue,
     filecolor=magenta,
     urlcolor=blue,
     citecolor=black
 }

\usepackage{rpm_SIunits}
\usepackage{rpm_acronyms}
\usepackage{rpm_math}
\usepackage{rpm_misc}

\usepackage{soul,color}
\usepackage{lipsum}

\usepackage{xcolor}

\DeclareMathOperator*{\argmin}{argmin}
\DeclareMathOperator*{\argmax}{argmax}

\usepackage[small]{caption}
\usepackage{adjustbox}
\usepackage{tikz} 

\title{\LARGE \bf
Camera Agnostic Two-Head Network for Ego-Lane Inference
}

\author{Chaehyeon Song${}^{1,2}$, Sungho Yoon${}^{1}$, Minhyeok Heo${}^{1}$, Ayoung Kim${}^{2}$ and Sujung Kim${}^{1\ast}$
\thanks{$^{1}$S. Yoon, M. Heo, S. Kim are with NAVER LABS, Seongnam-si, Gyeonggi-do, S. Korea {\tt\small [sungho.yoon, heo.minhyeok, sujung.susanna.kim]@naverlabs.com}}%
\thanks{$^{2}$C. Song and A.kim are with Department of Mechanical Engineering, SNU, Seoul, S. Korea {\tt\small [chaehyeon,ayoungk]@snu.ac.kr}}%
\thanks{$^\dagger$C. Song was with NAVER LABS as a research intern when conducting this work. This research was also supported by IITP grant No.2022-0-00480.}
\thanks{$^\ast$Corresponding author.}
}
\begin{document}

\maketitle
\thispagestyle{empty}
\pagestyle{empty}

\begin{abstract}

Vision-based ego-lane inference using \ac{HD} maps is essential in autonomous driving and advanced driver assistance systems. The traditional approach necessitates well-calibrated cameras, which confines variation of camera configuration, as the algorithm relies on intrinsic and extrinsic calibration. In this paper, we propose a learning-based ego-lane inference by directly estimating the ego-lane index from a single image. To enhance robust performance, our model incorporates the two-head structure inferring ego-lane in two perspectives simultaneously. Furthermore, we utilize an attention mechanism guided by vanishing point-and-line to adapt to changes in viewpoint without requiring accurate calibration. The high adaptability of our model was validated in diverse environments, devices, and camera mounting points and orientations. 
\end{abstract}

\section{Introduction}
Identifying the location of an ego vehicle on the road plays a fundamental role in providing relevant information to the user within advanced driver assistance systems. In order to operate with limited onboard sensors, visual localization is commonly employed combined with low-cost \ac{GNSS}. To extend the adoption of these systems to open road environments without assumptions about the camera, the robustness and installation flexibility have emerged as new challenges.

Conventional visual localization methods estimate~\cite{CVPR-2019-Sarlin,CVPR-2020-Sarlin, RAL-2021-Yoon,IROS-2022-Gao} the 6D pose using pre-built visual feature maps comprising millions of 3D points and their associated discriminative visual descriptors. While this method provides precise pose estimation, the substantial computational cost involved in extracting visual descriptors and the effort involved to create and maintain a prior map create entry barriers. To reduce these expensive burdens, other approaches~\cite{ICRA-2020-Asghar,RAL-2022-He, ICRA-2021-Wang} have explored the use of vectorized \ac{HD} maps as localization map sources. They extract road markers and lanes from the environment and matching them with elements of the maps. This process enables a significant reduction in computational resources and storage requirements. Alternative approaches \cite{IV-2015-Lee,ITSC-2017-Ballardini,IV-2019-Kasmi} go even further to maximize the cost-effectiveness by inferring the ego-lane in which an ego-vehicle is positioned, as it is more closely associated with the driving strategies, instead of estimating the absolute pose.
\begin{figure}[!t]
    \centering
    \includegraphics[trim=70 180 210 170,clip,width=1.0\columnwidth]{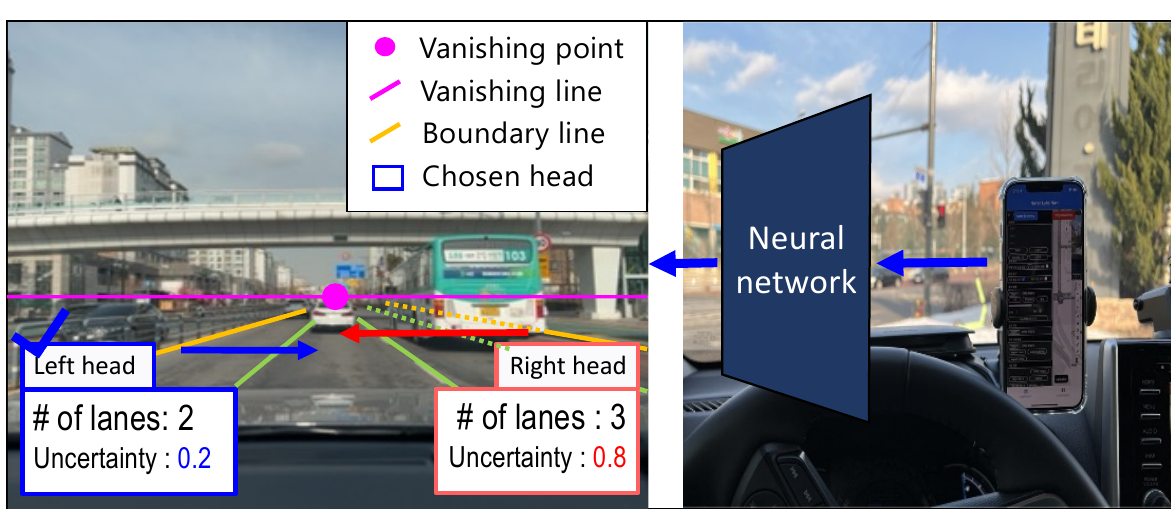}
    \caption{\textbf{Ego-lane inference operating on a smartphone.} Our research aims to develop a model that accurately determines the lane of the ego vehicle with a single image. Our model has two heads that predict the number of lanes from the ego-lane to the left and right boundary lines. Their uncertainty outputs are utilized to make a better decision between two heads. This model operates robustly even with images obtained by a mobile phone by leveraging vanishing points and lines to consider variations in mounting cameras.
    }
    \label{fig:overview}
    \vspace{-5mm}
\end{figure}

So far, these methods have primarily focused on cost-effectiveness, neglecting the flexibility of camera installations. These methods prioritize the extraction and analysis of road-related information, which requires an additional pre-processing step, such as bird-eye view projection and prior knowledge of intrinsic and extrinsic camera parameters. Another challenge is the complex open road environment, in which crucial lane information may not be observed on the camera, such as situations where the central line is not visible due to camera installation or when the lane markings are not fully covered on extremely wide roads.

In order to ensure applicability across diverse open roads with various camera settings, we introduced an end-to-end ego-lane inference network to identify the lane index in which the ego-vehicle is positioned as \figref{fig:overview}. To address the challenges posed by various situations, we proposed the integration of two novel components: 1) a two-head model with uncertainty and 2) an attention process reflecting vanishing point and line information. By incorporating these components, our approach presented remarkable adaptation in various environments and camera configurations. Also note that our proposed method does not rely on any specific localization map source by excluding \ac{HD} maps in the inference phase, making it applicable to a wide range of map types. The main contributions of the paper are summarized as follows:

\begin{itemize}
    \item 
    We present a single image-based ego-lane inference utilizing the deep neural network. This network operates in real-time without prior information about the camera and is validated in diverse environments and images captured by mobile phones.

    \item 
    The two-head structure, a novel approach of inferring ego-lane in two perspectives simultaneously, enhances the robustness in various scenarios and estimates uncertainty for selecting the more reliable outcome. 

    \item 
     The attention mechanism, with a VPL (vanishing point and line)-aware context vector, effectively deals with diverse scenarios such as variations in camera parameters, mounting configurations, and viewpoint differences.
\end{itemize}

\vspace{0.1cm}
\section{Related Works}
\label{sec:relatedwork}
\subsection{Visual Localization}
In recent visual localization work, structure-based localization methods have been widely studied using learning-based local features. 
When a coarse camera pose was provided by low-cost \ac{GNSS} or image retrieval methods \cite{PAMI-2018-Arandjelovic}, a camera pose could be estimated by the \ac{PnP} with \ac{RANSAC} after image matching \cite{CVPR-2019-Sarlin,CVPR-2020-Sarlin, RAL-2021-Yoon,IROS-2022-Gao}. 
Recent studies \cite{CVPR-2018-DeTone,NIPS-2020-Tyszkiewicz, CVPR-2022-Li} actively developed feature point detection and description for more accurate pose estimation. 
For robust matching, \citet{CVPR-2019-Sarlin} proposed the learning-based feature matching method using graph neural networks. While these methods provided precise 6D pose estimation, they come with the substantial computational cost involved in extracting local visual features and the effort required to create and maintain a prior map.

Other approaches \cite{IP-2016-Du,ICRA-2020-Asghar,IV-2013-Schreiber,RAL-2022-He, ICRA-2021-Wang, RAL-2020-Jeong} has proposed to utilize \ac{HD} maps as localization map sources. \ac{HD} maps provide compact representations that are crucial for driving scenarios while requiring less memory storage.

To deal with this issue, existing methods leveraged distinctive semantic features of road objects, such as lanes, road markers, and pole-like objects on the sidewalk. \citet{ICRA-2020-Asghar} involved a topological and \ac{ICP}-based lane map matching and implemented \ac{EKF}-based localization for multi-sensor fusion. He and Rajkumar~\cite{RAL-2022-He} proposed to use a spatial-temporal particle filter with a factor graph to resolve lane-matching ambiguity from lane-marker detection and a real road environment. While similar in objectives, our work differs from these approaches as we aim to estimate the vehicle's current lane directly from a single image, without relying on maps. The \ac{HD} map is only necessary for compensating for the lateral offset of low-cost \ac{GNSS}'s coarse positioning on roads. Consequently, due to limited reliance on \ac{HD} maps, our algorithm functions independently of the specific map types.

\begin{figure*}[t!]
    \centering
    \includegraphics[trim=50 185 19 115, clip,width= 0.98\textwidth]{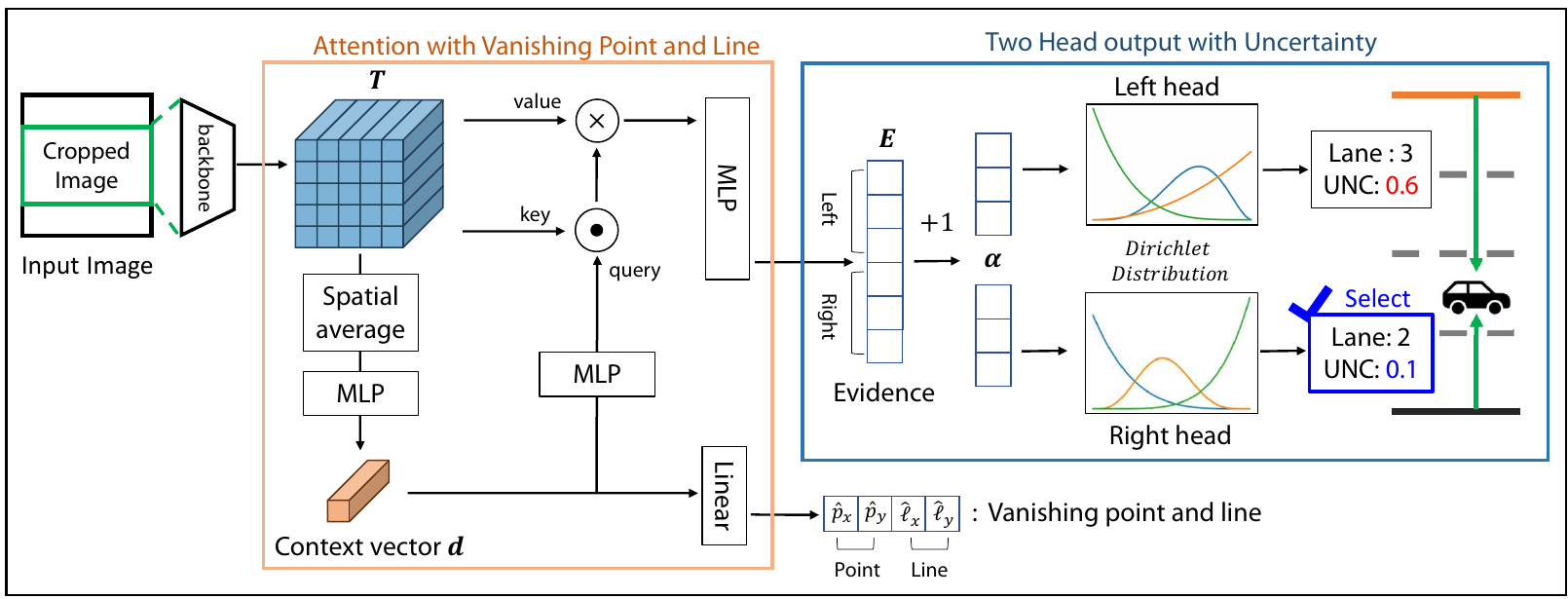}
    \caption{\textbf{The architecture of our model.}
    First, \ac{CNN} abstracts the local features from the input image. These local features are utilized in the attention process as key and value. The context vector from the average of the features and guided by the vanishing point and line becomes a query vector passing through one MLP layer. After going through this process, the model outputs the evidence vector $E$, and this vector is converted to probabilities and uncertainty of each head.
    }
    \label{fig:pipeline}
    \vspace{-2mm}
\end{figure*}

\begin{figure}[t!]
    \centering
    \includegraphics[trim=70 180 300 123,clip,width=1.0\columnwidth]{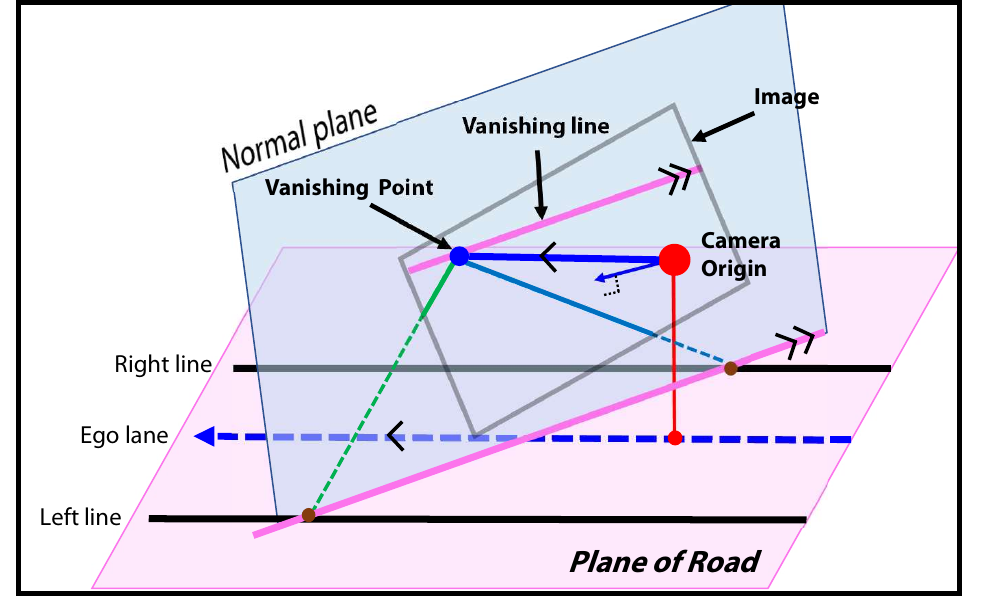}
    \caption{\textbf{Geometric relationship between \ac{VPL} and the ego-lane.} 
    With the premise of a planar road, the vanishing line is parallel to the plane of the road, and the lay from the camera origin and passing through the vanishing point is parallel to the lanes in the road.
    }
    \label{fig:geometry}
    \vspace{-2mm}
\end{figure}

\subsection{Ego-Lane Inference}
Various methods have been developed to estimate ego-lanes \cite{CVPR-2009-Gao} for lane-level navigation. Similar to \ac{HD} map-based localization methods, \citet{IV-2017-Rabe} exploited on-board sensors (e.g., odometry, \ac{GNSS}, camera, and radar) and \ac{HD} map \cite{IV-2014-Bender} as inputs of the particle filter. In recent years, \ac{BN} have also been utilized in probabilistic approaches \cite{IV-2015-Lee,ITSC-2017-Ballardini,IV-2019-Kasmi}. \citet{IV-2015-Lee} developed a mapless ego-lane recognition model considering various visual clues through pre-processing of bird-eye view images. To deal with the failure of detecting features from sensor input, a hidden Markov Model was utilized \cite{ITSC-2017-Ballardini} \cite{IV-2019-Kasmi}. \citet{IV-2019-Kasmi} extrapolated the lane information for the entire road by assuming each lane had the same shape and was located at an equal lateral interval in the road.
In contrast to the above works, we focused on developing end-to-end \ac{DNN} approach that eliminates intricate detection and matching steps as well as any pre-processing procedures involving camera parameters.

\vspace{0.1cm}
\section{Method}
\label{sec:method}

We propose an end-to-end ego-lane inference network that identifies the ego-lane in which the ego vehicle is positioned. To effectively handle diverse open road environments, we designed the architecture of our inference network that includes two key components, as depicted in \figref{fig:pipeline}. The first component is the VPL (vanishing point and line)-aware attention model that focuses on informative road features while considering geometric variations. The other component is the two head network output with uncertainty that enhances the overall performance by utilizing two complementary outcomes. Further details on these techniques are explained in the following sections.

\subsection{Attention with Vanishing Point and Line}

While CNNs excel at extracting local information such as line segments, they may not capture sufficient context for the ego-lane inference task. Determining the ego-lane requires considering the relative position of the ego-vehicle in relation to reference boundary lines. To make a model selectively focus on relevant boundary lines, our proposed model integrated an attention mechanism, which combines a global context vector with local information.

First, we passed an input image through the backbone network to generate the feature map $\boldsymbol{T}$, while maintaining a consistent aspect ratio by cropping the input image.
We spatially averaged the input feature map $\boldsymbol{T}$ and then pass it through a multi-layer perceptron (MLP) to obtain the context vector $\boldsymbol{d}$, which encodes global context information. 

The attention mechanism is similar to \cite{NIPS-2017-Vaswani} as
\begin{eqnarray*}
    \boldsymbol{Q} &=& \boldsymbol{L}_{q}\boldsymbol{d},\\ 
    \boldsymbol{K} &=& \boldsymbol{L}_{k}\boldsymbol{T}, \hspace{0.5cm}
    \boldsymbol{V} = \boldsymbol{L}_{v}\boldsymbol{T}, \\
    \boldsymbol{A} &=& \text{softmax}(\frac{\boldsymbol{Q}\boldsymbol{K}^T}{\sqrt{N}}),\\
\label{eq:evidence}
    \boldsymbol{E} &=& \max(0,\boldsymbol{L}_{2}\  \text{GELU}(\boldsymbol{L}_{1}\boldsymbol{A}\boldsymbol{V})).
\end{eqnarray*}
$\boldsymbol{L}_i$ is an linear layer and $N$ is the normalize factor. 
$\boldsymbol{Q}$, $\boldsymbol{K}$ and $\boldsymbol{V}$ correspond to the query, key and value of attention process respectively.
Finally, through the attention process, we obtain the output evidence vector $\boldsymbol{E}$, which will be further described at Section~\ref{sec:twohead}.

During the attention mechanism, the global context vector $\boldsymbol{d}$ captures the most informative context, such as relevant lines in our case. However, given the potential variations in camera setups, this contextual information can differ. To further enhance the robustness of the global context vector against such camera variations, we introduced a vanishing point and line loss as additional geometric guidance when computing the context vector.

The geometric relationship between vanishing point-and-line and the ego-lane are described in \figref{fig:geometry}. Specifically, the ray originating from the camera and passing through the vanishing point is parallel to the road lanes, and the vanishing line is parallel to the road plane. This configuration provides an intuitive understanding that the ego-lane generally intersects the vanishing point and forms a perpendicular relationship with the vanishing lines in the image. Note that these characteristics require some assumptions, such as planer roads and parallel lines. To alleviate these constraints, we gradually decay the weight of geometric loss as Eq. \eqref{eq:final loss}.

Based on these observations, we concluded that the vanishing points and lines are appropriate components for describing the geometric relationship between the road plane and the camera regardless of the camera configuration. 
Note that estimating the precise vanishing points and lines is a challenging problem and requires complicated architecture~\cite{NIPS-2019-Zhou}. 
Our objective is not to directly estimate the precise vanishing point and line themselves. Instead, our goal is to guide the context vector in capturing the geometric information. 

To do this, we pass the context vector $\boldsymbol{d}$ through a linear layer to obtain a 4-dimensional output representing the estimated vanishing point $\boldsymbol{\hat{p}}=(\hat{p}_x, \hat{p}_y)$ and line $\hat{\boldell}=(\hat{\ell}_x, \hat{\ell}_y)$. They were trained through the geometric loss function $\mathcal{L}_g$ with the ground-truth vanishing point $\boldsymbol{p}_{\text{gt}}$ and line $\boldell_{\text{gt}}$ as follows:
\begin{equation} 
    \label{g loss}
    \mathcal{L}_{g} = ||\boldsymbol{\hat{p}} - \boldsymbol{p}_{\text{gt}}||^2 + (1 - \hat{\boldell}\cdot{}\boldell_{\text{gt}}).
\end{equation}
\vspace{0.01cm}

The vanishing points and lines were trained through self-supervised learning, which did not require labeling the ground-truth value manually. 
The initial value of vanishing points and lines could be estimated using the \ac{HD} maps as \figref{fig:dataset} (a) and geometric characteristics described above. While applying homography transformation $\boldsymbol{H}$ to the input image, the vanishing points and lines alter according to the following rules as
\begin{equation*} 
    \label{homography}
    \boldsymbol{p}_f = \boldsymbol{H}\boldsymbol{p}_i, \hspace{0.8cm} \boldell_f = \boldsymbol{H}^{-T}\boldell_i.
\end{equation*}
\vspace{0.01cm}

\subsection{Two Head Outputs with Uncertainty} \label{sec:twohead}

Depending on camera configuration or road environments, it is not guaranteed that the central line is consistently visible in a single image. To address performance degradation in such situations, we have proposed a two-head model for robust and reliable ego-lane inference. Moreover, to fully leverage the advantages of the proposed two-head model, we also introduce the uncertainty of each head, empowering the algorithm to choose the more reliable outcome automatically.

For uncertainty, we leveraged the evidential deep learning theory~\cite{NIPS-2018-Sensoy}, which does not require additional inference time compare to other methods such as Bayesian Neural Networks and methods based on model ensembles.
The core of this theory is to design a model that estimates evidence of each class rather than direct class probability. 
The evidence values are converted as parameters of Dirichlet distribution, and the probability can be acquired by applying expectation at the probability distribution.

Let's simplify the problem by considering the ego-lane inference task as a $M$-class classification problem for each head.
Given the evidence $\boldsymbol{E}=\{e_m\}$ for each class $m=1, ..., M$,  Dirichlet parameter $\alpha_m$ corresponding to the $m^{th}$ class is defined as 
\begin{equation*}
 \alpha_m = e_m + 1,
\end{equation*}
then we computed the expected probability of the $m^{th}$ class $p_m$ and the total uncertainty of head $u$ as follows:
\begin{equation}
\label{eq:prob}
 p_m = \frac{\alpha_m}{\sum \alpha_m},  \hspace{0.2cm}  u = \frac{M}{\sum \alpha_m}.
\end{equation}
The uncertainty of head $u$ was utilized to assess the reliability of the output from each head, where lower uncertainty values indicate higher reliability.

Extending this to the two-head model with the left and right heads represented by the index $i\in\{\text{left}, \text{right}\}$, our method can identify the most reliable head $h$ by considering the uncertainty of the $i^{th}$ head, denoted as $u^i$.
\begin{equation*}
 h = \argmin_i u^i.
\end{equation*}
Therefore, the final output of the proposed two-head model is represented as the following set:
\begin{equation*}
\left( h, \argmax_m p^{h}_{m} \right)
\end{equation*}
where $p^{h}_{m}$ denotes the expected probability for the $m^{th}$ class in the $h^{th}$ head. 
In this study, we employed the two-head model for ego-lane inference from left and right references. 
The output can be interpreted as either the $m^{th}$ lane based on the leftmost side or the $m^{th}$ lane based on the rightmost side.

In order to train the evidence for the $m^{th}$ lane for each head $i$, similar to typical classification problems, we employ the Maximum Likelihood (ML) loss for the $i^{th}$ head as 
\begin{equation}
    \label{ML loss}
    \mathcal{L}_{ml}^i = -\sum_{m=1}^{M} y_{mi}\log(p_{mi}),
\end{equation}
where $y_{mi}$ and $p_{mi}$ represent the ground-truth label and the expected probability for the $m^{th}$ class of the $i^{th}$ head, as defined in Eq.~\eqref{eq:prob}.

Proposed two-head model can deal with the road environments where there are $M$ lanes on each side, resulting in a maximum $2M$ lanes.
Therefore, it can be possible that the ground-truth for each head exceeds $M$, when the ego-vehicle is attached to the left or right side. 
In such cases, the network ideally should output an uncertainty of 1. 
To achieve this, we adopt the Kullback-Leibler divergence (KLD) proposed by \cite{NIPS-2018-Sensoy} to penalize situations where the ground truth $y$ has value for indices larger than $M$.

\begin{equation} 
    \label{KLD loss}
    \mathcal{L}_{kld}^i = (1-\sum_{m=1}^{M} y^{i}_{m})\cdot{}
    KL[D\left(\boldsymbol{p}|\boldsymbol{\alpha}^{i}\right)||D\left(\boldsymbol{p}|\boldsymbol{1}\right)],
\end{equation}
where $D\left(\boldsymbol{p}|\boldsymbol{\alpha}^{i}\right)$ and $D\left(\boldsymbol{p}|\boldsymbol{1}\right)$ denote Dirichlet distribution with parameters $\boldsymbol{\alpha}^{i}=[\alpha^{i}_{1},...\alpha^{i}_{M}]$ and uniform Dirichlet distribution respectively~\cite{NIPS-2018-Sensoy}. 

The final loss is combined by Eq. \eqref{ML loss}, \eqref{KLD loss},  and \eqref{g loss} as 
\begin{equation}
    \label{eq:final loss}
    \mathcal{L}_{total} = \sum_{i}\left[\mathcal{L}^i_{ml}+ w\cdot{}\mathcal{L}^i_{kld}\right]+(1-w)\mathcal{L}_{g},
\end{equation}
where $w=\min(1, \frac{2*\text{iter}}{\text{maxiter}})$ is the weight to ensure stable convergence, gradually increasing during the training procedure. 

At the initial stages of training, our model assigns more significance to the geometric loss, $\mathcal{L}_{g}$. However, as training progresses, the importance of the evidence is progressively increased.

\section{Experiments}
\label{sec:experiment}
\subsection{Implementation Details}
\label{sec:implementation}

\textbf{Data preprocessing:} We maintain a consistent aspect ratio between the cropped image and resizing shape. This is especially crucial as road appearances can differ significantly between vertical and horizontal images if the image is only resized. We cropped the top and bottom evenly until the aspect ratio is 1:1.5 since the vanishing line is typically located at the center of the image. We then resized the image to $384\times256$ while ensuring that the cropped image and resizing shape match the aspect ratios.

\textbf{Model architecture:} The Resnet18 \cite{CVPR-2016-He} is utilized for the feature extraction process. We altered the down-sample ratio from 32 to 16 to reinforce the detecting performance of small objects; hence, the shape of the feature matrix $\boldsymbol{T}$ is $24\times16\times512$.
In the attention process, the number of attention heads is 8 and the head dimension is 64. 
And we set the number of classes of each head $M$ to 3 since there were no roads with more than six lanes in our study.

\begin{figure}[t!]
    \centering
    \includegraphics[trim=55 80 60 30,clip,width=1.0\columnwidth]{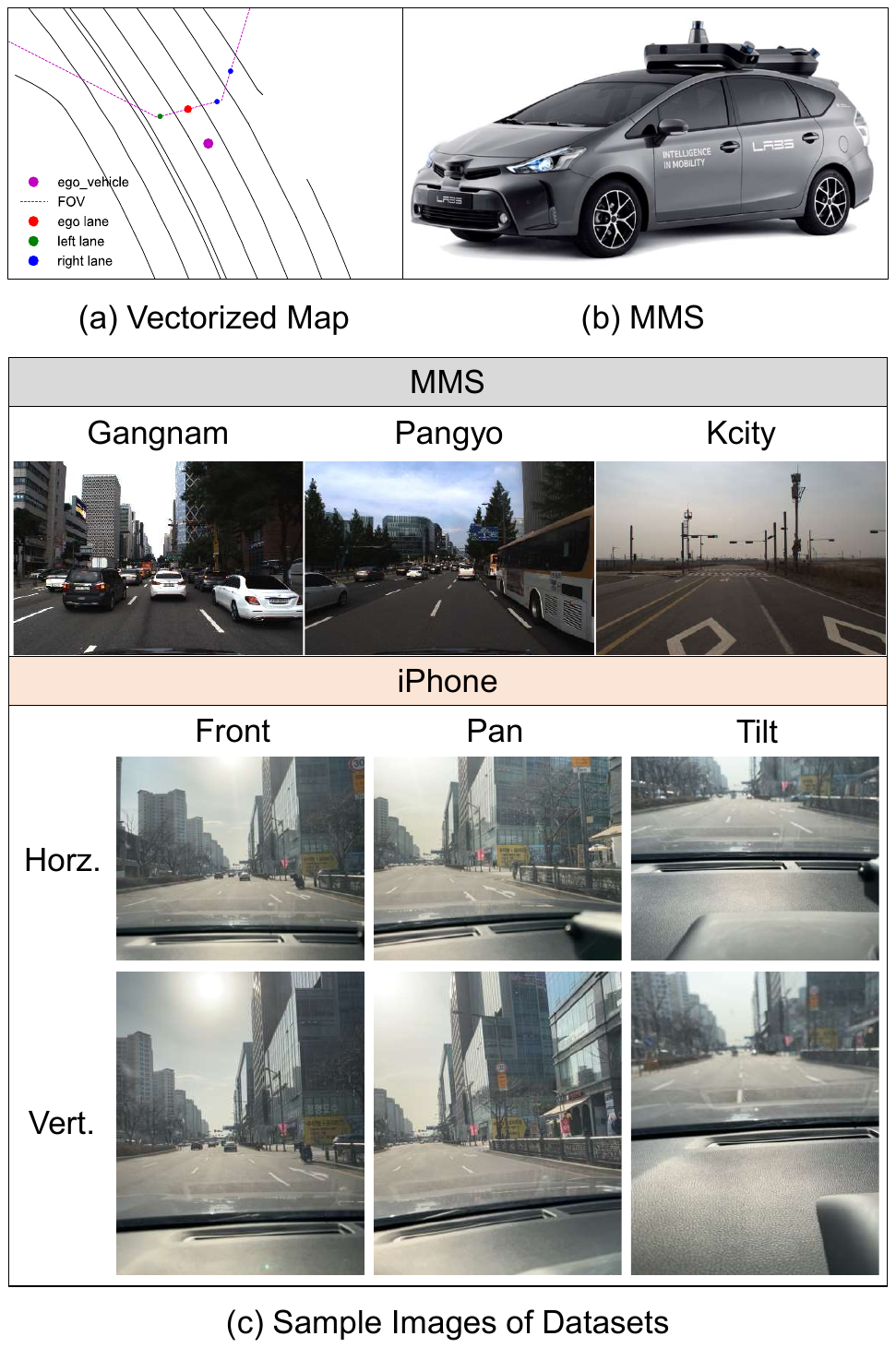}
    \caption{\textbf{Description of datasets.}
   Diverse datasets were generated, comprising RGB images captured by industrial cameras and iPhones, along with corresponding lane numbers from the boundary line using vectorized map and MMS. The field of view, depicted by the magenta line in the vector map, was used to determine the correct lane numbers in the image perspective. \texttt{Gangnam}, \texttt{Pangyo}, \texttt{Kcity} comprises images obtained by the MMS in various regions, and six iPhone datasets lie on different camera configurations.
    }
    \label{fig:dataset}
\end{figure}

\begin{table*}[t!]
\vspace{0.1cm}
\centering
\setlength{\tabcolsep}{10pt}
    \caption{\textbf{The F1 scores(\%) of different models.} 
    We investigated the effectiveness of each component by comparing the F1 scores. The results demonstrate that both the attention mechanism and the proposed VPL-aware loss enhance the performance across diverse environments and different camera setups. Despite the challenges such as limited view area, out-of-focus, and rotated images, our proposed model consistently achieved the highest F1 scores.}
    \begin{tabular}{c c| cccccccc}
        \toprule
            \multicolumn{2}{c}{Model}&\multicolumn{8}{c}{Dataset}\\
        \midrule
        \multirow{2}{*}{VPL}&\multirow{2}{*}{Attention}& \texttt{Gangnam} & \texttt{Pangyo} &\texttt{K-City}& \texttt{iPhone} & \texttt{iPhone} & \texttt{iPhone} & \texttt{iPhone} & \texttt{iPhone}\\ 
        &&MMS& MMS & MMS & Horz. & Vert. & Front & Pan & Tilt\\
        \midrule
        \midrule
            &&93.9 & 90.6 & {93.0} & 84.0 & 87.0 & 86.2 & 86.1 & 84.4 \\ 
            \checkmark&& 94.3 & 91.2 & {93.0} & 89.2 & {89.8} & 89.9 & {89.6} & {89.0}\\ 
            &\checkmark& {96.4} & {93.7} & \textbf{94.7} & {89.5} & {89.8} & {92.2} & 89.1 & 87.5 \\ 
            \checkmark&\checkmark& \textbf{97.0} & \textbf{93.9} & \textbf{94.7} & \textbf{91.0} & \textbf{91.4} &\textbf{92.3} & \textbf{89.8} & \textbf{91.6}  \\ 
        \bottomrule
    \end{tabular}
    \label{tab:F1scores}
\end{table*}

\textbf{Training:} In the training procedure, four augmentations(i.e. color jitter, homography transformation, masking, vertical flipping) are applied.
We determined the mini-batch to 512 and the learning rate to $5.0e^{-5}$. The Adam \cite{ArXiv-2017-Loshchilov} optimizer and cosine annealing scheduler with warm-up \cite{NIPS-2017-Vaswani} were used. 
The early stopping was executed since it was shown that the generalization performance decreased at more than 40 epochs.

\subsection{Dataset}
The existing relative dataset~\cite{ITSC-2017-Ballardini} is insufficient to test the robustness and generalization ability of our model since it consists of images captured under consistent highway environments with a fixed number of lanes. Furthermore, there is no dataset that focuses on the flexibility of camera installation. 

Therefore, we built several novel datasets(i.e. \texttt{Gangnam}, \texttt{Pangyo}, \texttt{K-City}, and \texttt{iPhone}) utilizing the vectorized road map and \ac{MMS} developed by an autonomous driving group in NAVER LABS as shown in \figref{fig:dataset} (a) and (b). Each dataset consists of images and their corresponding labels, which are the number of lanes from the ego lane to two(left and right) boundary lines. \texttt{K-City} comprises 3,712 images and represents a rural area, which is a specially designed place for the autonomous driving testbed in Korea. 
The last \texttt{iPhone} dataset consists of 1,563 mobile phone images of various camera poses(i.e. horizontal, vertical, pan, and tilt) obtained by iPhone pro 14 in the business district as shown in \figref{fig:dataset} (c).

To evaluate the generalization capacity of our model in different environments, we divided the \texttt{Gangnam} dataset into training and testing datasets. Then we only used the training dataset of \texttt{Gangnam} during the training phase and tested using other datasets, including testing datasets of \texttt{Gangnam}.

\subsection{Quantitative Evaluation}
\label{sec:Ablation}
\textbf{Model performance:} The inference times of our model are 59.6\ms~on an i9 CPU and 7.4\ms~on an RTX3070 GPU, which substantiates the lightweight and capacity to operate in real-time scenarios. We also verify the F1 scores of our model in various environments as shown in \tabref{tab:F1scores}. Since we trained our model only in the \texttt{Gangnam} dataset, we could evaluate the model performance with generalization ability by comparing the results in \texttt{Gangnam}, \texttt{Pangyo}, and \texttt{K-City} datasets. Although the place appearances differ from the training set in \texttt{Gangnam}, our model performed well, achieving over 94\% in F1-score. 
We analyzed how the model performances change in different mounting setups using the \texttt{iPhone} dataset, which has five mounting directions: horizontal, vertical, front, pan, and tilt. Despite adverse images resulting from limited view area, out of focus and rotations, our model achieves F1 scores over 90\%. The performance on \texttt{iPhone}(Pan) dataset is slightly lower than others since the left center line is not observed prevalently due to pan angles as shown in \figref{fig:dataset}~(c).
\begin{figure}[t]
    \centering
    \includegraphics[trim=152 125 170 135,clip,width=1\columnwidth]{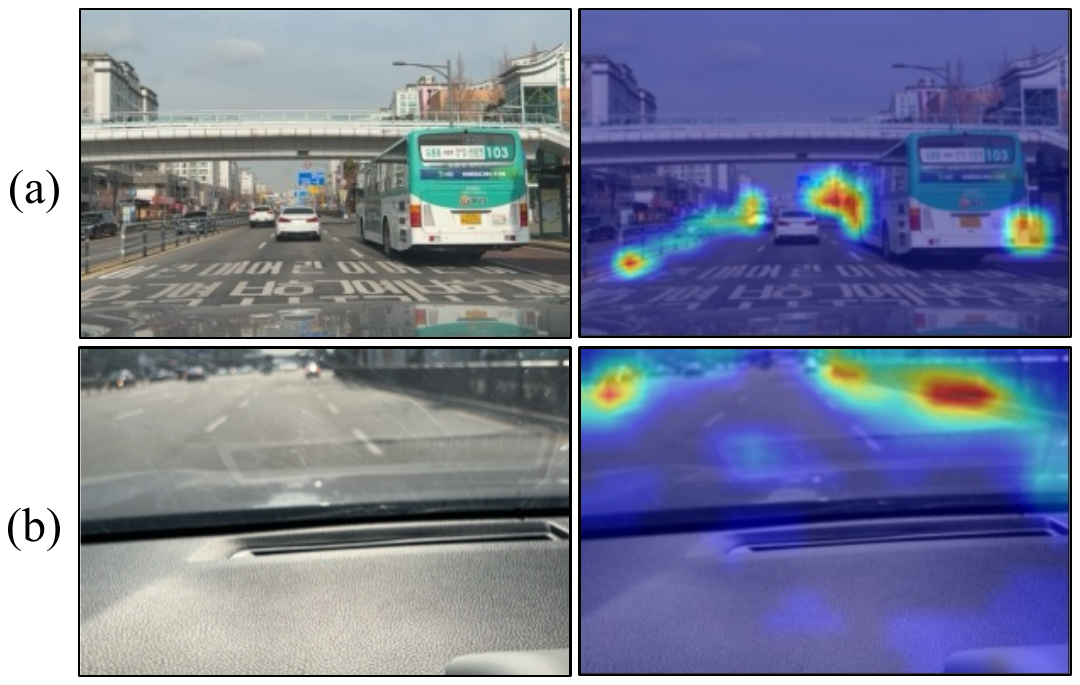}
    \caption{\textbf{Visualizing attention maps.} The attention layer effectively identifies reference boundary lines with attention to the \ac{VPL}-aware context vector despite the challenges such as (a) occlusions, and (b) low road visibility. 
    }
    \label{fig:attmap}
    \vspace{-5mm}
\end{figure}

\textbf{Ablation study:} 
The baseline, a \ac{CNN} combined with the two-head structure, worked fairly well in the datasets obtained by \ac{MMS} and achieved the 90.6\% F1 score, as seen at \tabref{tab:F1scores}. However, the performance dropped easily in the iPhone dataset by 84.4\% in the tilted image set. This is because the network was not designed to adjust installation changes. On the other hand, our method using \ac{VPL} guidance works robustly in diverse mounting configurations and  with up to 4.6\% enhancement in the tilt dataset, while the performances of \ac{MMS} datasets are similar to the baseline.
The guiding effect through the vanishing point and lines is further maximized when combined with the attention layer. By successfully integrating global and local information, the overall performance is improved by up to 2.6

\subsection{Attention Map Visualization}
We studied the attention map to observe how our model works, and we found that the attentions were mostly on finding left and right boundary lines. In \figref{fig:attmap}\,(a), our attention layer found the right boundary line well, although it was occluded and separated by a bus. Besides, \figref{fig:attmap}\,(b) shows one of the tilted images taken by a mobile phone. Because the tilted image focused on the deck board, the road appearance was unclear and difficult to perform a task. Our model captured the boundary line robustly, even in a difficult situation.

\subsection{Uncertainty Analysis}
In this section, we have investigated whether the tendency of our uncertainty coincides with general intuition in various driving circumstances.

\begin{figure}[t]
    \centering
    \includegraphics[trim=20 5 10 105,clip,width=1.0\columnwidth]{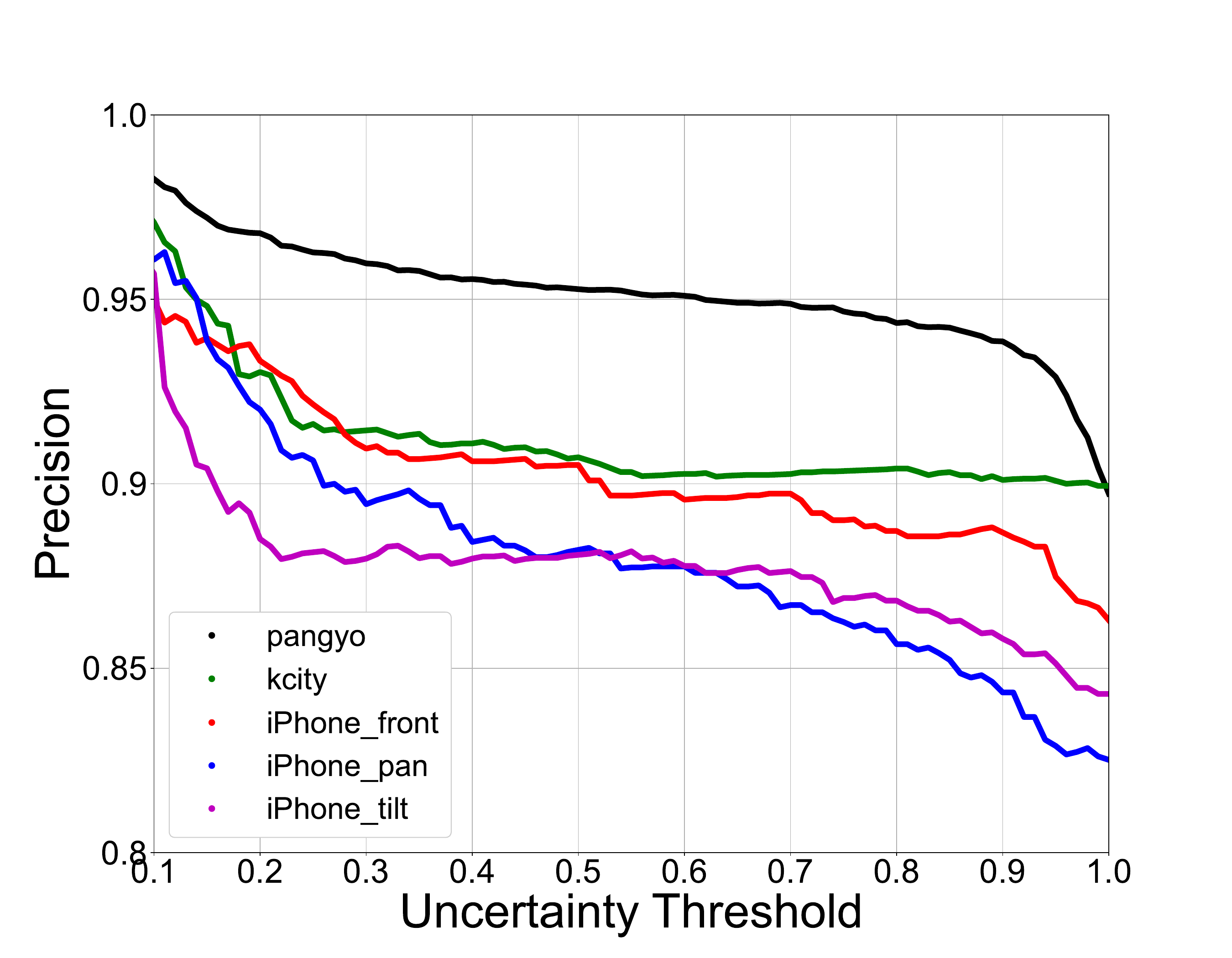}
    \caption{\textbf{Uncertainty-Precision curve}. The monotonic decreases graph supports the uncertainties are well designed and enables users to decide the precision level by controlling the threshold of uncertainty.}
    \label{fig:UP_curve}
\end{figure}

\textbf{Rejection based on uncertainty:} The uncertainty is not only used for selecting a more reliable head but also for rejecting the final output itself. If the uncertainty of the selected head exceeds a threshold determined by the user, a user could not trust the result and takes a different strategy such as assuming that it is in the same lane as the previous time step. By adjusting the thresholds, the user could obtain the desired precision of the model.
\figref{fig:UP_curve} represents the variation in precision based on different threshold values. The monotonic decrease in the graph for all datasets indicates that the designed uncertainty aligns well with general intuition. When the thresholds are above 0.9, a sharp decline in precision is observed, which is mainly associated with situations where lane detection solely based on images is impossible, such as intersections.

\begin{figure}[t]
    \centering
    \includegraphics[trim=15 20 30 80,clip,width=0.89\columnwidth]{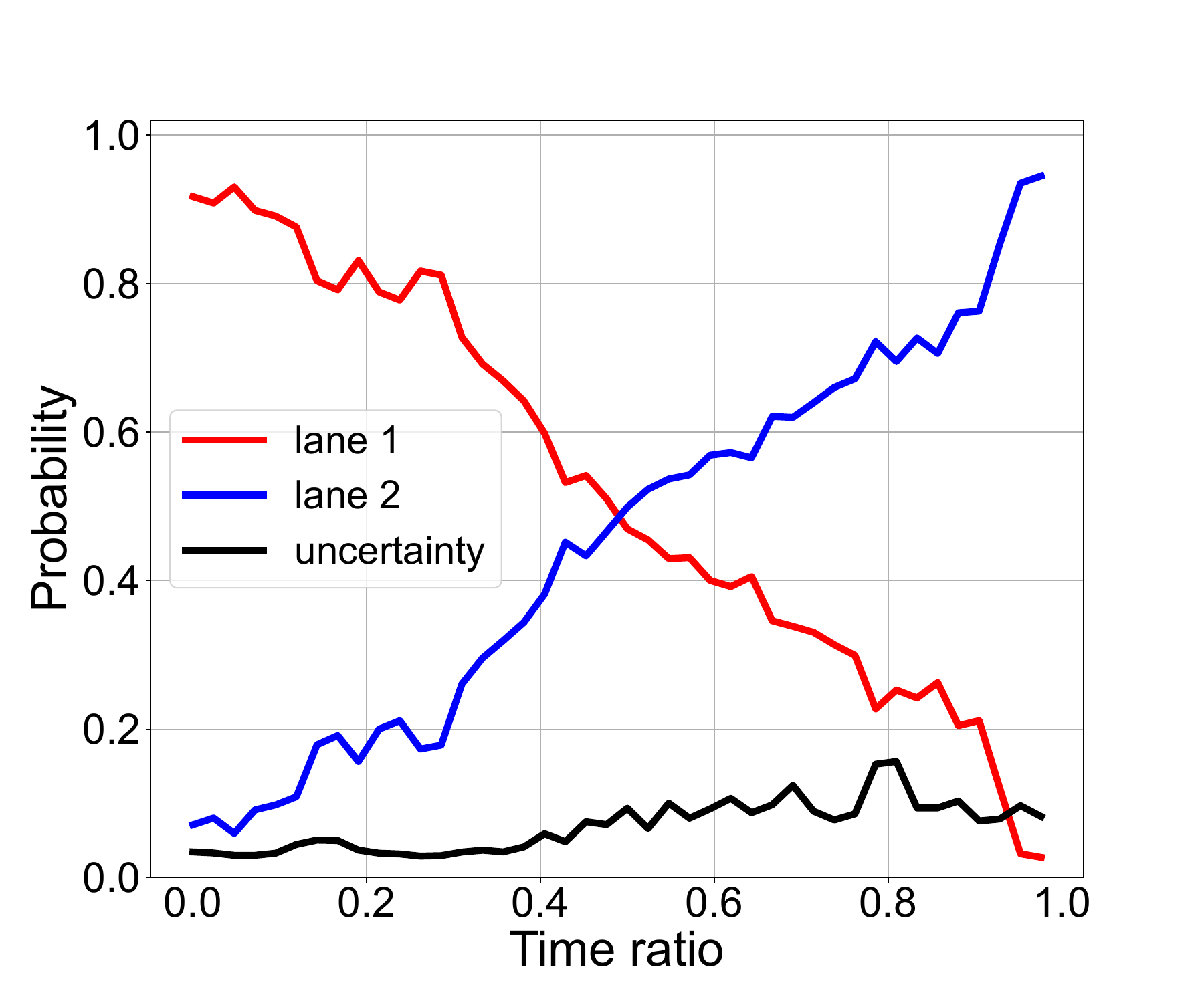}
    \caption{\textbf{Probabilities and uncertainty during lane changing.} When the ego lane alters from the first to the second lane, the probability of class 0 and class 1 are gradually reversed. Meanwhile, the uncertainty does not increase since the model is confident about the result.}
    \label{fig:lane_changing}
    \vspace{0.4cm}
\end{figure}

\textbf{Lane changing:} \figref{fig:lane_changing} substantiate that our model generates reliable probability and uncertainty while changing lanes.  When the ego lane alters from the first to the second lane, the probabilities of lane 1 and lane 2 cross in the middle. The uncertainty is not growing up considerably during this moment.
This implies that the uncertainty is not directly associated with the entropy of a probability distribution.
\begin{figure}[t]
    \centering
    \includegraphics[trim=130 150 235 160,clip,width=0.99\columnwidth]{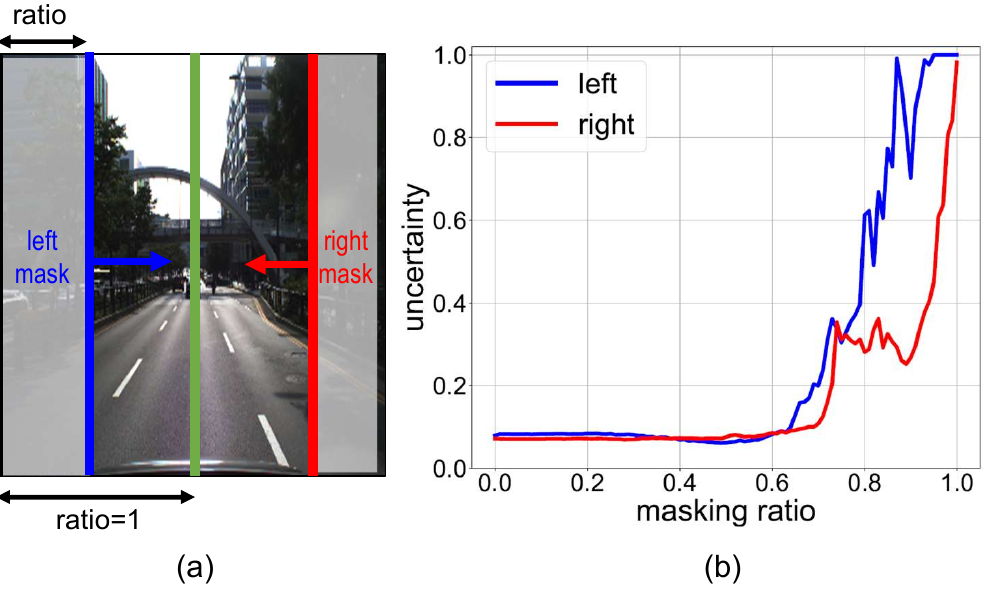}
    \caption{\textbf{Uncertainty variation under occlusions.} (a) Occlusion masks. (b) The graph describes the uncertainty of our model, aligned with human intuition. The uncertainty does not increase until the masking ratio of 0.5 since our model captures even a small clue at the boundary line. }
    \label{fig:unc_variation}
\end{figure}

\textbf{Occlusion:} We analyzed how the uncertainty outputs are changing in our model by reducing the visible area in an input image. To define the region of interest, we introduced an occlusion mask that begins at the edges of the image and ends at the vanishing point. As shown in \figref{fig:unc_variation}, the level of uncertainty remains stable until the occlusion ratio reaches 0.5. However, beyond this point, the uncertainty increases significantly by the ratio reaching up to 0.7. Overall, our results highlight the model's ability to infer decision uncertainty based on the visibility of a boundary line, particularly as the visible area in the image decreases. It represents that our model outputs reliability properly in the limited visibility.

\begin{figure}[t]
    \centering
    \includegraphics[trim=105 230 325 110,clip,width=0.99\columnwidth]{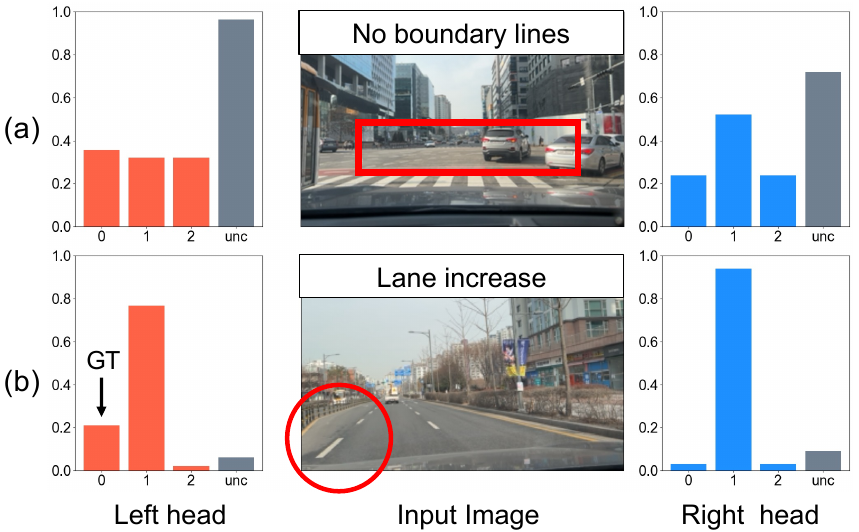}
    \caption{\textbf{The limitations on our model}. (a) If boundary lines are not visible in the image, our model gives up with high uncertainty. (b) When the maximum lane alters, the model may confuse. 
    \vspace{0.5cm}  
    }
    \label{fig:failure}
    \vspace{-5mm}
\end{figure}

\subsection{Limitation}
\figref{fig:failure} depicts the typical limitations that our model cannot estimate an ego lane. Specifically, \figref{fig:failure}\,(a) represents the limited scenario at the crossway. When the boundary line is not visible explicitly, the model does not infer the ego lane, resulting in high uncertainty. 
In \figref{fig:failure}\,(b), humans can deduce the ground truth of the left head is zero as the ego lane diverges to two lanes, although the left boundary line has not yet shifted. However, the model only makes a decision based on a specific position in the input image, which may lead to false outcomes.

\subsection{Application on Lane-level Navigation}
\begin{figure}[t]
    \centering
    \includegraphics[trim=120 185 430 175 ,clip,width=0.85\columnwidth]{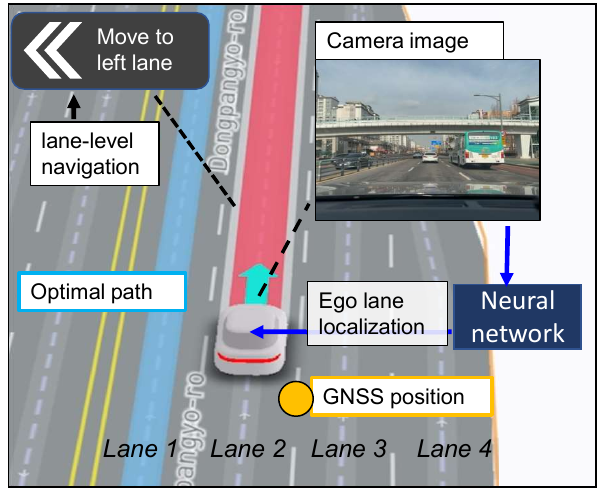}
    \caption{\textbf{The application on to a lane-level navigation}. By using our model, improved navigation that provides lane-level information is possible even on smartphones}    
    \label{fig:application}
\end{figure}

The coarse \ac{GNSS} position can be compensated in the lateral direction using the lane number of an ego vehicle in a \ac{HD} map as \figref{fig:application}. With a smartphone and our model, the advanced navigation system could guide how many lanes to move rather than just left or right turns. The brief demonstration video is accessible through the following link \url{http://tinyurl.com/ELITHNdemo}.

\section{Conclusion}
\label{sec:conclusion}
We propose an ego-lane inference network to identify the lane index in which the ego-vehicle is positioned using a single image. The incorporation of a two-head network with uncertainties enhances the overall performance, overcoming limited view area.
Our proposed VPL-aware attention mechanism also effectively handles diverse scenarios such as variations in camera parameters, mounting configurations, and viewpoint changes. This adaptability makes our approach well-suited for wide-ranging camera setups including mobile phones. To validate the model's effectiveness, we extensively evaluated its performance across various road environments and camera mountings, demonstrating its robustness and efficacy.

\balance
\small
\bibliographystyle{IEEEtranN} 
\bibliography{string-short,references}

\end{document}